\documentclass[letterpaper]{article} 
\usepackage{aaai2027}  
\nocopyright
\usepackage[hyphens]{url}  
\usepackage{graphicx} 
\usepackage{natbib}  
\usepackage{caption} 
\usepackage{algorithm}
\usepackage{algorithmic}
\usepackage{amsmath}
\usepackage{amssymb}
\usepackage{newfloat}
\usepackage{listings}
\usepackage[table]{xcolor}

\DeclareCaptionStyle{ruled}{labelfont=normalfont,labelsep=colon,strut=off} 
\floatstyle{ruled}
\newfloat{listing}{tb}{lst}{}
\floatname{listing}{Listing}
\usepackage{booktabs}

\title{Decoupled Self-Forcing Distillation for Streaming Talking Head Generation}
\author{
    Yanru An$^1$,
    Ruiyan Wang$^1$,
    Wenwu Wei$^1$,
    Rui Bu$^2$,
    Qi Wang$^2$,
    Hongwei Hu$^2$
    \\
    Zhengxue Cheng$^1$,
    Rong Xie$^1$,
    Li Song$^1$,
    Wenjun Zhang$^1$
}
\affiliations{
    $^1$Shanghai Jiao Tong University
    $^2$Ant Group\\
}

\begin{document}

\maketitle

\begin{abstract}
Streaming talking-head generation produces each frame as its driving audio arrives, yet fidelity and efficiency have so far pulled in opposite directions: end-to-end methods condition a video diffusion model on audio directly and achieve high quality but only at large scale, while cheaper two-stage methods generate an intermediate motion representation and trail in fidelity. We argue the cost of the former lies in the target of fusion: the video latent is dominated by identity, appearance and background, none of which audio bears on, so coupling audio to every pixel blurs detail and wastes capacity. We instead fuse conditions in a low-dimensional identity-disentangled motion space, routing audio and motion captions by their temporal granularity, and generate motion latents with a small causal autoregressive transformer that a pretrained diffusion renderer turns into video. Conditions thus control video transitively, and high fidelity no longer requires a large backbone. Streaming this decomposition needs both models to be causal, and the exposure-bias problem could be solved by self-forcing given a bidirectional teacher. But there is no such teacher in motion space. Our decoupled self-forcing distillation resolves both models under one frozen teacher: conditioned on motion, it distills the renderer into a block-causal student; unconditionally, it scores rendered rollouts against real videos, supervising motion by the video it produces. This lifts the fidelity ceiling from the motion generator onto the stronger renderer. The two models run as parallel causal streams, reaching 15.4 FPS at 1.3 s latency with no quality degradation.
\end{abstract}

\section{Introduction}
\label{sec:intro}

Audio-driven talking-head generation has advanced rapidly with video diffusion models (VDMs)~\cite{hallo,echomimic,omnihuman,hunyuanvideoavatar}, yet real-time applications like live streaming demand \emph{streaming} synthesis, in which each frame is generated at low latency as its driving audio arrives. Existing methods fall into two families. \textbf{End-to-end} methods finetune a pretrained VDM~\cite{wan,cogvideox} with cross-attention for condition injection~\cite{liveavatar,streamavatar,rest,avatarforcing} and use autoregressive (AR) distillation~\cite{causvid,dmd,sf} to make it few-step and block-causal. \textbf{Two-stage} methods instead generate a compact motion representation from audio and hand pixels to a small renderer~\cite{artalk,dystream,ditto,posetalk}. The two trade against each other: end-to-end methods reach the highest fidelity but introduces a large number of parameters, while two-stage methods have lower computational costs but trail in fidelity~\cite{rest}.

We trace this tension to \emph{where} conditions are aligned. End-to-end methods align audio to the video latent, and blame the resulting cost on the difficulty of multimodal fusion. We argue the difficulty lies not in the fusion operation but in its target: the video latent is dominated by identity, texture and lighting conditions, none of which is causally determined by audio. Coupling audio to every pixel forces it to interact with texture, lighting and identity; trained at scale, the model regresses these entangled factors toward their conditional mean, blurring fine detail much as a mean-seeking objective would. The natural target is not pixels but head motion, a low-dimensional, identity-free manifold where alignment is leakage-free and cheap. We therefore build on the two-stage paradigm. Its known weakness is a fidelity gap: existing methods optimize the motion generator entirely within motion space, isolated from the renderer that consumes its output, so fidelity is capped by the motion generator rather than the far stronger renderer.

We propose \textbf{Motar}, a two-stage streaming talking-head generation method. We adopt the identity-disentangled motion latent of X-NeMo~\cite{xnemo} and generate it with a causal transformer whose per-frame diffusion head~\cite{mar} models continuous values without the quantization error of a discrete codebook~\cite{artalk}. A hierarchical conditioning scheme fuses audio and a motion caption by a Q-Former~\cite{blip2} into a global coarse condition, while per-frame audio drives a local branch for lip synchronization in the $d$-dimensional motion space, keeping it lightweight. The key advantage is indirect but decisive: motion controls video and audio controls motion, so conditions control video transitively. High fidelity thus needs no large backbone; the renderer need not know what \emph{angry} means, only what a motion latent looks like in pixels.

Streaming this decomposition exposes two exposure-bias problems: the AR motion model drifts on its own rollout, and the bidirectional renderer must be distilled into a few-step block-causal one~\cite{causvid}. Self-forcing~\cite{sf} is the standard remedy, but presupposes a frozen bidirectional teacher, which exists in video space and not in motion space. Our \emph{decoupled self-forcing distillation} resolves both under the same frozen X-NeMo teacher: its conditional form distills the renderer via distribution matching~\cite{dmd}; its unconditional form scores rendered rollouts against real videos, backpropagating through the frozen renderer into the motion model. Motion is thus optimized by its video-space output rather than its latent-space precision, lifting the fidelity ceiling onto the renderer and countering the diversity collapse of regression-based self-forcing.

Our contributions are three-fold:
\begin{itemize}
\item We relocate multimodal fusion to a low-dimensional motion manifold, giving transitive control over video that decouples fidelity from backbone scale.
\item We design a causal motion generator with a continuous diffusion head and hierarchical conditioning separating coarse text--audio control from frame-level lip sync.
\item We propose \emph{decoupled self-forcing distillation}, resolving two exposure-bias problems under one frozen teacher and lifting the two-stage fidelity ceiling onto the renderer, yielding a fully streaming two-stream causal pipeline.
\end{itemize}

\section{Related Work}

\subsection{Audio-Driven Talking Head Generation}
\label{sec:rw-thg}
Early methods targeted lip synchronization~\cite{wav2lip}, and were later extended toward expressive faces and natural head motion~\cite{sadtalker,audio2head}. Diffusion-based approaches now split along two axes. \emph{End-to-end} methods fine-tune a pretrained VDM to attend to audio directly~\cite{hallo,echomimic,emo,omnihuman,memo,sonic,hunyuanvideoavatar}, achieving strong lip-sync and naturalness. \emph{Two-stage} methods instead generate a compact identity-agnostic motion representation from audio and delegate pixels to a renderer, using either explicit keypoints~\cite{ditto,liveportrait} or learned disentangled latents~\cite{anitalker,vasa1,float,imtalker}, cutting cost by orders of magnitude. The standard objection to the latter family is that it is bottlenecked by the fidelity of the generated motion, and so trails end-to-end methods in naturalness~\cite{rest}. We attribute this to how the decomposition is trained: existing two-stage methods optimize the motion generator entirely within motion space, in isolation from the renderer that consumes its output. We close this gap by supervising the motion generator with a video-level distribution matching objective.

\subsection{Streaming and Real-Time Avatars}
\label{sec:rw-streaming}
Two routes lead to real time. One distills a large bidirectional VDM into a causal, few-step student~\cite{talkingmachines,liveavatar,streamavatar,rest,avatarforcing}, which streams but only at large-parameter scale and, in several cases, across multiple GPUs; READ~\cite{read} accelerates the backbone yet remains non-autoregressive and cannot stream at all. The other makes the \emph{motion} generator autoregressive, which is lightweight, but couples it to a renderer whose expressiveness is structurally bounded --- either an explicit 3D avatar driven by parametric coefficients~\cite{artalk,gagavatar}, whose expressions are confined to the span of a fixed basis, or a warping-based decoder~\cite{dystream,lia}, which transports pixels frame by frame and struggles with fine detail and temporal consistency. Either way, only one end of the pipeline is temporally causal. We pair an implicit motion latent with a diffusion renderer and causalize \emph{both}, running them as two parallel streams.

\subsection{Autoregressive Diffusion Distillation}
\label{sec:rw-distill}
AR models trained with teacher forcing degrade over long rollouts, as inference exposes them to their own compounding error. Diffusion Forcing~\cite{df} conditions at arbitrary noise levels; CausVid~\cite{causvid} distills a block-causal VDM with DMD~\cite{dmd}, but still accumulates error over long horizons; Self-Forcing~\cite{sf} closes the gap directly, unrolling the student on its own predictions during training and matching its output distribution to a frozen bidirectional teacher, with subsequent work pushing this to minute-scale~\cite{longlive,causalforcing}. Every method in this line presupposes that such a teacher exists. The premise holds in video latent space, where large pretrained bidirectional diffusion models are available, but fails in motion latent space, where no pretrained bidirectional motion teacher exists and an AR motion generator therefore has exposure bias with no score to correct it against. We resolve both cases under a single frozen video teacher.

\begin{figure*}[t]
\centering
\includegraphics[width=0.95\textwidth]{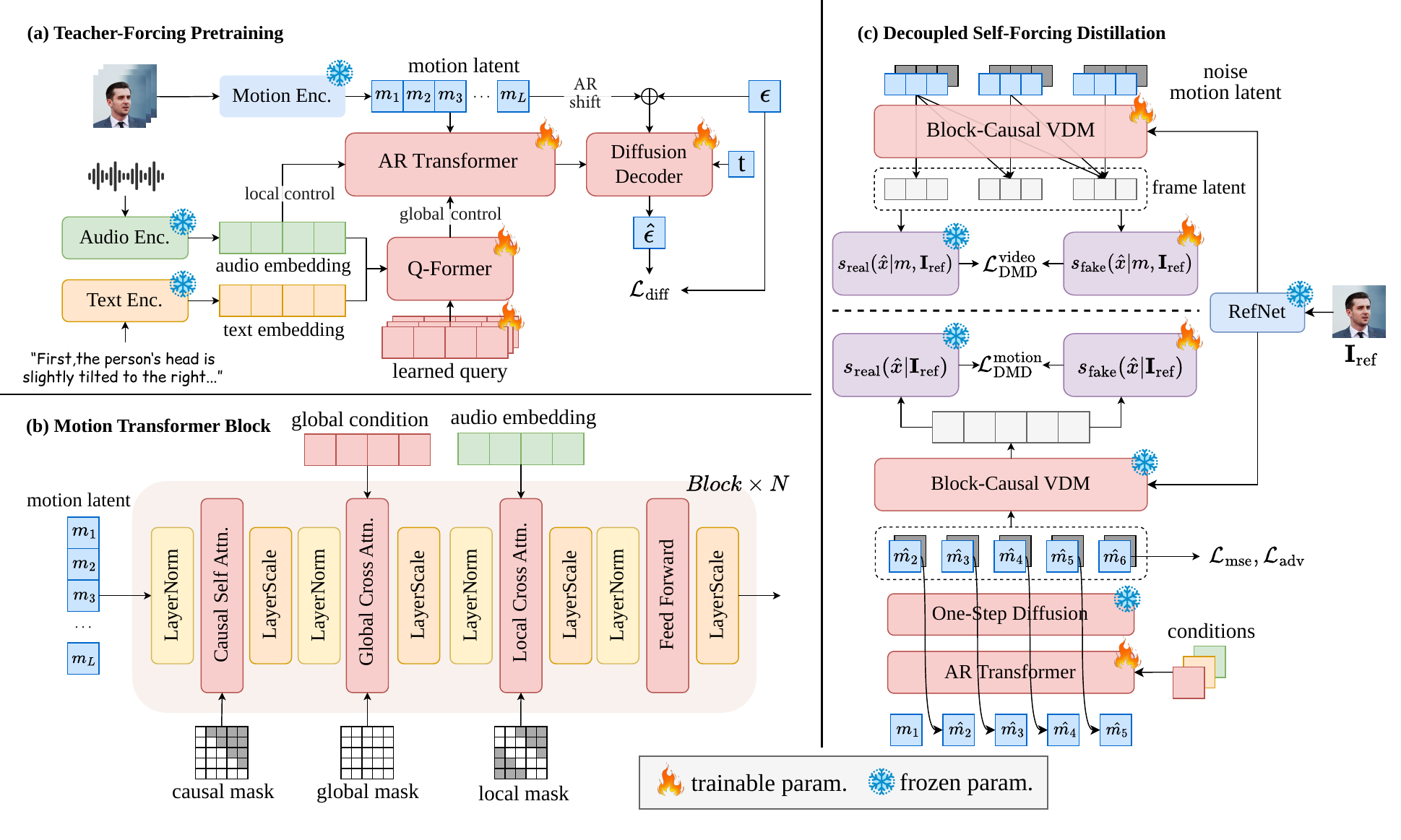}
\caption{Overview of Motar. (a) An AR motion transformer models causal dependencies over motion latents, with an efficient diffusion head producing continuous-valued latents; a user-written motion caption and the driving audio are fused into a global condition query by a Q-Former projector. (b) Hierarchical conditioning: the global query is injected by full cross-attention for coarse, sequence-level control, while raw audio embeddings are injected by windowed cross-attention for frame-level lip articulation. (c) Decoupled self-forcing distillation: the frozen bidirectional teacher $G$ supervises both branches by distribution matching. Conditioned on motion, it distills $G$ into a block-causal student $G_\phi$; unconditionally, it matches the rendered motion rollout against the distribution of real talking videos.}
\label{fig:method}
\end{figure*}

\section{Method}

The overview of our proposed Motar is shown in Fig.~\ref{fig:method}. Given a reference image $\mathbf{I}_{\text{ref}}$, a driving audio $c_a$, and an optional motion caption $c_t$ describing the desired emotion, amplitude and head movement, our goal is to synthesize a talking-head video in which the subject speaks and emotes accordingly while preserving the appearance of $\mathbf{I}_{\text{ref}}$.

\noindent\textbf{Motion--rendering decomposition.}
Following Sec.~\ref{sec:intro}, we align conditions with motion rather than pixels. We adopt the pretrained identity-disentangled motion encoder $\mathcal{E}_{m}$ of X-NeMo~\cite{xnemo}, which maps a video $\mathbf{V}=(\mathbf{I}_1,\dots,\mathbf{I}_T)$ to a 1D motion latent sequence $\mathbf{M}=(\mathbf{m}_1,\dots,\mathbf{m}_T)$, $\mathbf{m}_t\in\mathbb{R}^{d}$, paired with a diffusion decoder $G$ that reconstructs the video from motion latents and a reference frame, $G(\mathbf{M},\mathbf{I}_{\text{ref}})\approx\mathbf{V}$. Since $\mathcal{E}_m$ is identity-disentangled, $\mathbf{M}$ retains only the low-dimensional dynamics of expression and head motion, to which $c_a$ and $c_t$ are causally related. The task splits into
\begin{equation}
\mathbf{M}=\mathcal{F}_{\text{motion}}(c_a,c_t), \qquad \mathbf{V}=G(\mathbf{M},\mathbf{I}_{\text{ref}}),
\label{eq:decomp}
\end{equation}
where the first prediction task is low-dimensional and identity-free, needing only a small model (Sec.~\ref{sec:motion-backbone}), and the second is handled by $G$, distilled into a causal student in Sec.~\ref{sec:decoupled-sf}. Since $G$ decodes latents locally, the two run as parallel streams, rendering frame $t$ as soon as $\mathbf{m}_t$ is generated.

\subsection{Autoregressive Motion Latent Generation}
\label{sec:motion-backbone}

$\mathcal{F}_{\text{motion}}$ predicts $\mathbf{m}_t\in\mathbb{R}^d$ from the history $\mathbf{m}_{<t}$ and the conditions. As $\mathbf{m}_t$ is continuous rather than discrete, we follow MAR~\cite{mar} and replace the softmax head with a small conditional diffusion head $v_\phi$, avoiding the quantization error imposed by a VQ codebook on facial motion~\cite{artalk} and separating causal sequence modeling (the backbone $f_\theta$) from distribution modeling ($v_\phi$). With $\mathbf{z}_t = f_\theta(\mathbf{m}_{<t},\mathbf{c}^{g},\mathbf{c}^{l}_{t})$ the conditioning vector from the backbone, the head is trained with a $v$-prediction objective,
\begin{equation}
\mathcal{L}_{\text{diff}}=\mathbb{E}_{\tau,\boldsymbol\epsilon}\bigl[\|v_\phi(\mathbf{m}_t^{\tau},\tau,\mathbf{z}_t)-\mathbf{v}^{\text{gt}}\|_2^2\bigr],
\label{eq:diffloss}
\end{equation}
where $\tau$ is the diffusion timestep and $\mathbf{m}_t^{\tau}$ the noised target. The backbone consists of $N$ causal transformer blocks with RoPE\cite{rope}, unrolled at inference with a rolling KV cache; we cache pre-RoPE projections and re-apply RoPE relative to the current window, keeping training and streaming inference consistent.

\noindent\textbf{Hierarchical conditioning.}
How the two modalities enter $f_\theta$ is dictated by their temporal granularity. Audio is tied to lip movement frame by frame, meaningful only at frame resolution; a motion caption asserts a sequence-level property with no frame-level timestamp. Routing both through one cross-attention pathway would force a frame-level signal and a sequence-level one to share attention weights. We therefore split conditioning into two branches. The \emph{global} branch fuses text embeddings $\mathbf{e}^{\text{txt}}$ and audio embeddings $\mathbf{e}^{\text{aud}}$ through a Q-Former~\cite{blip2} with learnable queries into a compact sequence $\mathbf{c}^{g}=g_\psi(\mathbf{e}^{\text{txt}},\mathbf{e}^{\text{aud}})$ that every frame attends to; sharing these queries across the sequence biases it toward coarse, sequence-level control such as emotion and amplitude. The \emph{local} branch lets frame $t$ attend only to a short audio window $\mathbf{c}^{l}_t=\mathbf{e}^{\text{aud}}_{t-w_a:t+w_a}$ with window size $w_a=2$, supplying the resolution for lip sync; applied independently per frame, it needs no cross-frame cache. Each block applies causal self-attention, global and local cross-attention, and an FFN, each pre-normalized and gated by a learnable residual scale~\cite{cait}. Because fusion happens in the $d$-dimensional motion space, $g_\psi$ and the cross-attention layers cost only a fraction of the parameters such modules need on a video backbone. The backbone is pretrained with teacher forcing (TF), conditioning $\mathbf{z}_t$ on ground-truth history.

\noindent\textbf{Bounded look-ahead for streaming.}
Neither branch is strictly causal as the global query summarizes a window of audio and the local branch peeks $w_a$ frames ahead, yet both preserve streaming with bounded latency since their look-ahead is bounded and independent of sequence length. In practice the global branch is fed a sliding window rather than the whole utterance: audio arrives faster than motion is generated, so this window is available without stalling, and the caption is compressed by the Q-Former into an embedding carrying no fine-grained future content. The local peek is a deliberate concession: a phoneme spans several frames, so a window sharpens lip sync at a fixed, small delay. The model is thus not strictly frame-causal but has bounded look-ahead, keeping first-frame latency constant and inference indefinitely streamable while improving quality.

\subsection{Self-Forcing of the Motion Generator}
\label{sec:sf-motion}

At inference the model conditions on its own predictions, so any error in $\hat{\mathbf{m}}_t$ compounds. Closing this gap requires fine-tuning on the model's own rollout, which must be differentiable end to end. Backpropagating through a multi-step denoiser at every AR step is computationally prohibitive, as the graph grows with the product of sequence length and denoising steps. We therefore first distill $v_\phi$ into a single-step consistency sampler $v_\phi^{\text{one-step}}$~\cite{lcm}, which both enables real-time generation and makes the rollout affordable. During rollout, the ground-truth frame is replaced by the model's own prediction with probability $p_{\text{sf}}$, annealed $0\!\rightarrow\!1$, with $v_\phi^{\text{one-step}}$ frozen so gradients flow into $f_\theta$ alone; as $p_{\text{sf}}\!\to\!1$ the backbone is directly exposed to its own compounding error.

\noindent\textbf{Why regression alone fails.}
The natural supervision is a masked regression against ground truth,
\begin{equation}
\mathcal{L}_{\text{mse}}=\tfrac{1}{|\mathcal{V}|}\textstyle\sum_{t\in\mathcal{V}}\|\hat{\mathbf{m}}_t-\mathbf{m}^{\text{gt}}_t\|_2^2,
\label{eq:sf-mse}
\end{equation}
where $\mathcal{V}$ indexes rolled-out frames. But Eq.~\ref{eq:sf-mse} is mean-seeking: it drives the model toward the conditional expectation of plausible next frames rather than a sample from their distribution, collapsing motion amplitude and diversity over a long rollout.

\noindent\textbf{Adversarial loss in motion space.}
We therefore add an adversarial term~\cite{gan} matching the rollout to the real motion distribution rather than its mean. A discriminator $D$ judges whether an entire sequence is plausible, with no per-frame target to collapse onto:
\begin{equation}
\mathcal{L}_{\text{adv}} = \mathbb{E}_{\mathbf{m}^{\text{gt}}}[\log D(\mathbf{m}^{\text{gt}})] + \mathbb{E}_{\hat{\mathbf{m}}}[\log (1 - D(\hat{\mathbf{m}}))].
\label{eq:gan-loss}
\end{equation}
$D$ operates on motion sequences unconditionally, penalizing rollouts whose temporal statistics degenerate. $\mathcal{L}_{\text{mse}}$ is kept as a weak anchor, and $w_{\text{adv}}$ ramped in once $D$ warms up. This constrains motion within its own space; whether it renders into a plausible video is addressed in Sec.~\ref{sec:decoupled-sf}.

\subsection{Decoupled Self-Forcing Distillation}
\label{sec:decoupled-sf}

Exposure bias is standardly addressed by distilling a causal student against a bidirectional teacher. In our decomposition, $\mathcal{F}_{\text{motion}}$ is autoregressive but has no bidirectional teacher in its own space, while $G$ is such a teacher but is bidirectional and multi-step and must be made block-causal and few-step. We resolve both by Self-Forcing against the same frozen bidirectional $G$; the two branches differ in one respect: whether the score model is conditioned on motion.

\noindent\textbf{Branch 1: distilling a causal renderer.}
We convert $G$'s temporal attention from bidirectional to block-causal with a rolling KV cache over past blocks, leaving per-frame spatial and reference-conditioning layers unchanged. The student $G_\phi$ is unrolled block by block on its own blocks as at inference, and supervised by DMD: a frozen $G$ is the real-score model, and an online copy $s_{\text{fake}}$, updated by a denoising loss on the student's output, tracks its distribution. Both are conditioned on ground-truth motion $\mathbf{m}$, as the task is to reproduce the teacher's rendering of a given motion:
\begin{equation}
\nabla_{\hat{\mathbf{x}}_0}\mathcal{L}^{\text{video}}_{\text{DMD}}\propto s_{\text{fake}}(\hat{\mathbf{x}}_\tau\mid\mathbf{m},\mathbf{I}_{\text{ref}})-s_{\text{real}}(\hat{\mathbf{x}}_\tau\mid\mathbf{m},\mathbf{I}_{\text{ref}}),
\label{eq:dmd-video}
\end{equation}
where $\hat{\mathbf{x}}_0=G_\phi(\mathbf{x}_\tau,\tau,\mathbf{m},\mathbf{I}_{\text{ref}})$ and $\hat{\mathbf{x}}_\tau$ its renoised counterpart. The gradient approaches zero only when $G_\phi$'s distribution matches $G$'s. $G_\phi$ is then frozen for all subsequent training.

\noindent\textbf{Branch 2: matching motion in video space.}
A motion sequence is meaningful only insofar as it renders into a plausible video, yet Sec.~\ref{sec:sf-motion} supervises it entirely within motion space --- the fidelity gap two-stage methods are criticized for. With a differentiable renderer, we instead supervise motion by the video it produces: we decode a window of the rollout $\hat{\mathbf{m}}_{k:k+w}$ through the frozen $G_\phi$ into $\hat{\mathbf{x}}_0$ and apply DMD,
\begin{equation}\nabla_{\hat{\mathbf{x}}_0}\mathcal{L}^{\text{motion}}_{\text{DMD}}\propto s_{\text{fake}}(\hat{\mathbf{x}}_\tau\mid\mathbf{I}_{\text{ref}})-s_{\text{real}}(\hat{\mathbf{x}}_\tau\mid\mathbf{I}_{\text{ref}}),\label{eq:dmd-motion}\end{equation}\begin{equation}\frac{\partial\mathcal{L}^{\text{motion}}_{\text{DMD}}}{\partial\theta}=\frac{\partial\mathcal{L}^{\text{motion}}_{\text{DMD}}}{\partial\hat{\mathbf{x}}_0}\cdot\frac{\partial\hat{\mathbf{x}}_0}{\partial\hat{\mathbf{m}}}\cdot\frac{\partial\hat{\mathbf{m}}}{\partial\theta},\label{eq:chain}\end{equation}
where $\theta$ denotes $\mathcal{F}_{\text{motion}}$'s parameters. Since $G_\phi$ and $\mathbf{I}_{\text{ref}}$ are fixed, motion is the only free variable, and the video-level gradient is transported back onto it.

\noindent\textbf{Why this branch must be unconditional.}Conditioning the score models on the generated $\hat{\mathbf{m}}$, as in Eq.~\ref{eq:dmd-video}, fails because $G$ is a faithful conditional renderer: given any motion it renders a consistent video and assigns it high density. The two scores then agree, the gradient vanishes, and no force acts on motion. The failure is structural: $s_{\text{real}}(\cdot\mid\hat{\mathbf{m}})$ asks whether $\hat{\mathbf{x}}$ is a plausible rendering of $\hat{\mathbf{m}}$, and the answer is always yes, never whether $\hat{\mathbf{m}}$ is itself plausible. Conditioning on ground-truth motion is no remedy either: the video would be evaluated under a condition it was not rendered from, reintroducing a per-frame target and the mean-seeking behaviour of Eq.~\ref{eq:sf-mse}. We therefore drop the motion condition, obtaining an unconditional score model by fine-tuning $G$ on real videos with the motion condition removed, retaining only $\mathbf{I}_{\text{ref}}$, so that $s_{\text{real}}(\cdot\mid\mathbf{I}_{\text{ref}})$ models the marginal distribution of real talking videos. A collapsed motion now renders to a near-static video, which this model assigns low density, and the gradient is non-zero.

\noindent\textbf{Complementary supervision in two spaces.}
The full objective for $\mathcal{F}_{\text{motion}}$ combines three terms,
\begin{equation}
\mathcal{L}_{\text{motion}}=\mathcal{L}_{\text{mse}}+w_{\text{adv}}\mathcal{L}_{\text{adv}}+w_{\text{dmd}}\mathcal{L}^{\text{motion}}_{\text{DMD}},
\label{eq:motion-total}
\end{equation}
where $\mathcal{L}_{\text{mse}}$ is an anchor that keeps $\hat{\mathbf{m}}$ in the region where the score models are reliable before the distribution-level terms dominate. The other two are both distribution-matching but act in different spaces. Applying $\mathcal{L}_{\text{adv}}$ directly to the motion latents is computationally inexpensive and allows the full rollout to be evaluated jointly, thereby capturing long-range temporal statistics such as amplitude and rhythm. $\mathcal{L}^{\text{motion}}_{\text{DMD}}$ operates on the rendered video, evaluating exactly what the viewer sees and penalizing motion that is well-formed yet renders poorly, but observes only a short window per step since rendering is expensive. Supervising in both spaces covers what either alone would miss, lifting the fidelity ceiling: motion is no longer optimized in latent space alone but by the video it renders into, so system quality is bounded by the renderer $G_\phi$ rather than the motion generator $\mathcal{F}_{\text{motion}}$.

\section{Experiments}

\begin{table*}[t]
\centering
\resizebox{\textwidth}{!}{
\begin{tabular}{l|ccccccccc}
\toprule
\textbf{Method} &
\textbf{FID}$\downarrow$ &
\textbf{FVD}$\downarrow$ &
\textbf{CSIM}$\uparrow$ &
\textbf{E-FID}$\downarrow$ &
\textbf{$\Delta$Sync-C}$\downarrow$ &
\textbf{$\Delta$Sync-D}$\downarrow$ &
\textbf{FPS}$\uparrow$ &
\textbf{Lat.(s)}$\downarrow$ &
\textbf{Param.}\\
\midrule
SadTalker &
101.6 / 83.9 &
346.3 / 519.1 &
0.869 / \underline{0.883} &
0.372 / 0.425 &
3.39 / \underline{0.43} &
3.57 / 0.42 &
\textbf{23.8} &
\textbf{0.21} &
0.2B \\

AniPortrait &
\underline{43.5} / \textbf{38.5} &
\underline{216.5} / \underline{385.8} &
\textbf{0.904} / \textbf{0.901} &
0.108 / 0.258 &
1.56 / 1.82 &
1.66 / 1.73 &
0.83 &
242 &
1.3B \\

Hallo3 &
68.5 / 42.3 &
239.5 / 395.0 &
0.839 / 0.773 &
\textbf{0.095} / \textbf{0.189} &
4.28 / 0.76 &
3.01 / \underline{0.26} &
0.31 &
157 &
8.9B\\

StableAvatar &
46.1 / 45.8 &
514.8 / 662.3 &
0.801 / 0.771 &
0.294 / 0.261 &
\underline{1.09} / 2.35 &
\underline{0.56} / 2.80 &
0.86 &
230 &
1.7B \\

LiveAvatar &
\textbf{31.9} / 40.7 &
242.8 / 443.1 &
0.812 / 0.806 &
0.511 / 0.389 &
3.92 / 0.67 &
2.79 / \textbf{0.01} &
0.38 &
43.3 &
16.3B \\

AvatarForcing &
46.7 / 42.8 &
351.7 / \textbf{340.5} &
0.768 / 0.728 &
0.240 / \underline{0.192} &
2.66 / 1.34 &
2.13 / 1.77 &
14.8 &
\underline{0.61} &
1.4B \\

\midrule
Motar(Ours) &
45.0 / 50.5 &
\textbf{185.9} / 424.2 &
\underline{0.894} / 0.831 &
\underline{0.109} / 0.280 &
\textbf{0.08} / \textbf{0.06} &
\textbf{0.42} / 0.37 &
\underline{15.4} &
1.39 &
77M + 1.7B\\

\rowcolor{gray!12}
$G(\mathbf{M}^{\text{gt}},\mathbf{I}_{\text{ref}})$ &
44.2 / 50.2 &
185.1 / 467.6 &
0.899 / 0.766 &
0.097 / 0.292 &
0.11 / 0.45 &
0.28 / 0.53 &
0.82 &
247 &
1.7B
\\
\bottomrule

\end{tabular}
}

\caption{Quantitative comparison on MEAD and Hallo3 public subsets. Results are reported as MEAD / Hallo3. Best results are bolded and second-best results are underlined. The GT Sync-C is 1.68 in MEAD and 4.93 in Hallo3-data. The GT Sync-D is 12.22 in MEAD and 8.90 in Hallo3-data. $\Delta$Sync-C and $\Delta$Sync-D denote the absolute difference relative to GT. Param. denotes the parameter count of the backbone. Oracle $G(\mathbf{M}^{\text{gt}},\mathbf{I}_{\text{ref}})$ only serves as a reference and do not participate in comparison.}
\label{tab:quantitative}
\end{table*}

\subsection{Experimental Setup}

\noindent\textbf{Dataset.}
We combine MEAD~\cite{mead} and Hallo3-data~\cite{hallo3}, covering both acted emotional and in-the-wild talking-head videos. All clips are resampled to 25 fps and cropped to $512\times512$. We use Qwen2.5-VL-7B~\cite{qwen} to generate a textual caption for each clip, describing the speaker's emotion and head pose movement; caption accuracy was verified by manual inspection on a random subset.

\noindent\textbf{Implementation Details.}
We extract 512-D motion latents using the pretrained X-NeMo~\cite{xnemo} encoder, with text and audio features encoded by umT5-base~\cite{t5} and wav2vec2-base~\cite{wav2vec}, respectively. The motion generator comprises 8 causal transformer blocks (dim 512) paired with a 2-layer MLP diffusion head, and consists of only \textbf{77M} parameters. Training follows a four-stage schedule: (1) teacher-forced pre-training of the backbone, (2) distillation of the diffusion head into a single-step consistency sampler, (3) self-forcing fine-tuning unrolled over 256 frames (transitioning from $\mathcal{L}_{\text{mse}}$ to $\mathcal{L}_{\text{adv}}$), and (4) decoupled self-forcing distillation where the motion generator is supervised by the distilled block-causal renderer $G_\phi$ via $\mathcal{L}^{\text{motion}}_{\text{DMD}}$ (rendering a 64-frame window). $G_\phi$ is distilled following Self-Forcing, using 64-length rolling KV-cache, 8-frame denoising blocks and 4-step sampling. All training is conducted with Adam optimizer on 4 NVIDIA A100 GPUs. More details are provided in the Supplementary Material.

\noindent\textbf{Baselines.}
We compare our model against state-of-the-art talking head generation methods, including two-stage approaches (SadTalker~\cite{sadtalker} and AniPortrait~\cite{aniportrait}) and end-to-end frameworks (Hallo3~\cite{hallo3}, StableAvatar~\cite{stableavatar}, AvatarForcing~\cite{avatarforcing}, and LiveAvatar~\cite{liveavatar}). We additionally report the reconstruction oracle, formulated as $G(\mathbf{M}^{\text{gt}},\mathbf{I}_{\text{ref}})$, which decodes ground-truth motion latents and serves as an upper bound for any method based on this latent decomposition.

\noindent\textbf{Metrics.}
We evaluate visual quality using FID and FVD, and measure identity preservation with CSIM~\cite{arcface}. Lip synchronization is assessed via $\Delta\text{Sync-C}$ and $\Delta\text{Sync-D}$~\cite{syncnet}, which denote the absolute differences between the computed sync scores and the ground-truth values of the test set. Emotion controllability is quantified by E-FID~\cite{emo}, computed as the Fr\'echet Distance of expression parameters extracted following~\cite{efid}. Finally, computational efficiency is evaluated in terms of first-frame latency (Lat.), steady-state throughput and parameter count (Param.) with each method generating a 200-frames video with BF16 precision on a single NVIDIA H200 GPU.

\subsection{Quantitative Comparison}
Table~\ref{tab:quantitative} shows that our method leads decisively on lip synchronization: $\Delta$Sync-C and $\Delta$Sync-D are the best on both subsets, on par with the reconstruction oracle. Since the oracle is driven by GT motion, matching it means almost no synchronization error originates in our motion generator.

SadTalker achieves the highest FPS and the lowest first-frame latency as a lightweight CNN-based method, at the cost of significantly inferior visual quality. Our method runs at 15.4 FPS, faster than all diffusion-based baselines. While AvatarForcing exhibits lower first-frame latency, it underperforms our method across most quality metrics. The additional latency in our method mainly comes from the Q-Former fusion module and the generation of the first block of motion latent, and is further reduced in the audio-only setting where the fusion module is omitted. Despite this overhead, the first-frame latency is only around one second, making it suited for real-time interactive applications.

On visual quality, FID, FVD and E-FID are competitive but do not lead, most visibly on the Hallo3 subset. This is expected: our renderer is built on Stable Diffusion(SD) backbone, weaker than the DiT-based backbones of the end-to-end baselines, so absolute visual fidelity is capped below theirs. The crucial comparison is not against those baselines but against the oracle, which uses the teacher renderer. Metrics of Ours are close to $G(\mathbf{M}^{\text{gt}},\mathbf{I}_{\text{ref}})$, and where we slightly exceed it is attributed to the oracle being trained on a different dataset. This proximity is the intended result showing the motion generator is not the bottleneck, and that our method raises the capability ceiling to the renderer's, which the ablations in Sec.~\ref{sec:ablation} confirm directly. A stronger renderer would lift visual fidelity without any change to the motion side.

\subsection{Qualitative Comparison}

\begin{figure*}[t]
\centering
\includegraphics[width=0.9\textwidth]{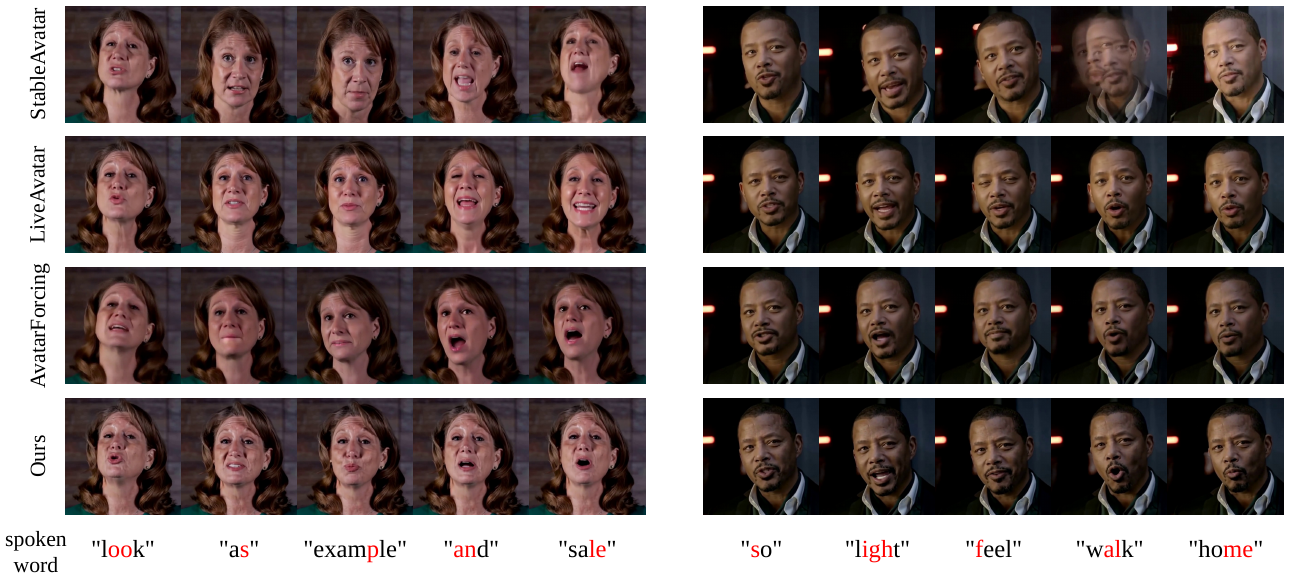}
\caption{Qualitative comparison with Wan-based end-to-end methods. For each result we show five frames with the spoken word beneath; the phoneme being articulated is highlighted in red (e.g.\ the ``oo'' in ``look''), so that lip shape can be checked against the sound at each frame. Our method produces the tightest audio--lip alignment and the sharpest facial detail.}
\label{fig:qualitative}
\end{figure*}

Figure~\ref{fig:qualitative} compares our method against StableAvatar, LiveAvatar and AvatarForcing, all built on the Wan backbone\cite{wan}, with the articulated phoneme marked in red at each frame. Two differences are consistent across examples.

First, lip synchronization. Our lip shapes track the highlighted phoneme frame by frame, whereas the baselines drift slightly. This matches the quantitative gap in $\Delta$Sync, and follows from the hierarchical conditioning: the local branch ties each frame directly to a short audio window, giving the frame-level resolution that a single global pathway cannot.

Second, facial detail. The end-to-end baselines couple audio to the full video latent through cross-attention, so the conditioning signal interacts with every pixel with texture and lighting included, neither of which audio should govern. Trained at scale, the model regresses these entangled factors toward their conditional mean, blurring fine detail much as a mean-seeking objective would. Our method never exposes these factors to audio since conditioning acts only on the identity- and appearance-free motion latent, while all pixel-level detail is carried by the reference image through the frozen renderer. Fine texture is thus preserved almost verbatim and stays stable as the head moves.

These two properties are why our method matches the perceptual quality of these larger DiT-based baselines despite a weaker backbone and faster inference: by restricting audio to motion, it spends its capacity where the conditioning signal belongs, rather than defending appearance against it.

\subsection{Ablation Study}
\label{sec:ablation}
In this section, we conduct ablation study on our proposed methods to demonstrate their effectiveness. More details and results are provided in the Supplementary Material.

\noindent\textbf{Training recipe.}
We ablate the training recipe of $\mathcal{F}_{\text{motion}}$ on 256-frame rollouts, isolating each term of Eq.~\ref{eq:motion-total}. Besides MSE, frame-wise cosine similarity (CosSim), and Fr\'echet motion distance (FMD) in latent space, we report two rollout-level ratios to ground truth, both ideally equal to $1$: Std-R, the ratio of the generated sequence's temporal standard deviation to GT, capturing motion amplitude; and VC-R(velocity consistency ratio), the same ratio on frame-to-frame differences, capturing motion pace.

\begin{table}[t]
\centering
\setlength{\tabcolsep}{4pt}
\begin{tabular}{@{}l|ccccc@{}}
\toprule
\textbf{Variant} & \textbf{MSE}$\downarrow$ & \textbf{CosSim}$\uparrow$ & \textbf{FMD}$\downarrow$ & \textbf{Std-R}$\uparrow$ & \textbf{VC-R}$\uparrow$ \\
\midrule
TF(50 steps)         & 0.518 & 0.637 & 30.33 & 1.001 & 1.208 \\
CM(1 step)           & 0.595 & 0.696 & 44.21 & 0.831 & 0.435 \\
\midrule
SF w/$\mathcal{L_\text{mse}}$        & \textbf{0.259} & \textbf{0.840} & 29.93 & 0.494 & 0.433 \\
+$\mathcal{L_\text{adv}}$       & 0.263 & 0.839 & 18.21 & 0.590 & 0.542 \\
+$\mathcal{L_\text{DMD}^\text{motion}}$ & 0.298 & 0.826 & \textbf{17.07} & \textbf{0.704} & \textbf{0.644} \\
\bottomrule
\end{tabular}
\caption{Ablation on the motion generator training recipe. We only compare Std-R and VC-R on SF variants as rollouts of TF variants collapse.}
\label{tab:ablation}
\end{table}

The two variants without self-forcing, the teacher-forced model (TF) and its consistency-distilled one-step version (CM) never see their own rollout in training, so error compounds at inference and MSE degrades. Their rollout-level ratios are unreliable as quality signals: TF's Std-R of $1.001$ and VC-R of $1.208$ do not indicate faithful dynamics but severe drift, whose erratic motion happens to inflate these ratios; CM instead collapses toward static output. We therefore read Std-R and VC-R only among the self-forced variants.

Self-forcing with $\mathcal{L}_{\text{mse}}$ removes exposure bias and sharply improves MSE, CosSim and FMD, but its regression objective is mean-seeking: the rollout collapses toward static, under-animated motion, giving the lowest Std-R ($0.494$) and VC-R ($0.433$) among SF variants. Adding $\mathcal{L}_{\text{adv}}$ evaluates the rollout's distribution rather than a per-frame target, restoring amplitude and pace and halving FMD. Adding $\mathcal{L}_{\text{DMD}}^{\text{motion}}$ improves FMD, Std-R and VC-R further as it is the only term supervising motion by the video it renders into. Both distribution-matching terms slightly raise MSE and CosSim, but this is by design rather than a cost: MSE rewards regressing toward the per-frame mean, precisely the collapse we aim to escape, so recovering amplitude and diversity necessarily moves MSE away from its minimum. FMD, Std-R and VC-R, which measure distributional rather than point-wise fidelity, are the metrics aligned with our goal, and all three improve.

\begin{figure}[ht]
\centering
\begin{minipage}{0.49\linewidth}    
\centering    
\includegraphics[width=\linewidth]{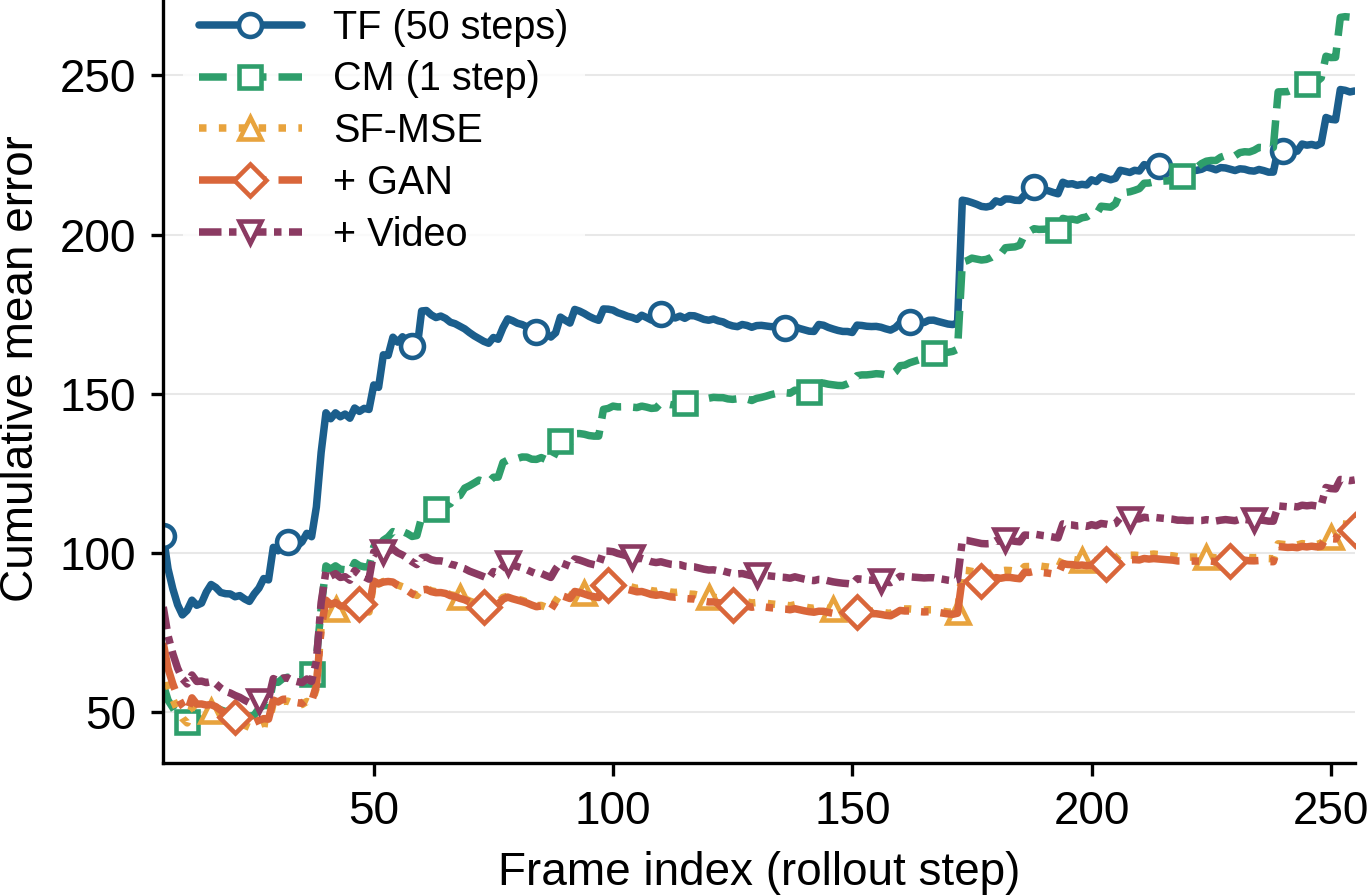}    
\caption*{(a) Cumulative error.}
\end{minipage}
\hfill
\begin{minipage}{0.49\linewidth}    
\centering    
\includegraphics[width=\linewidth]{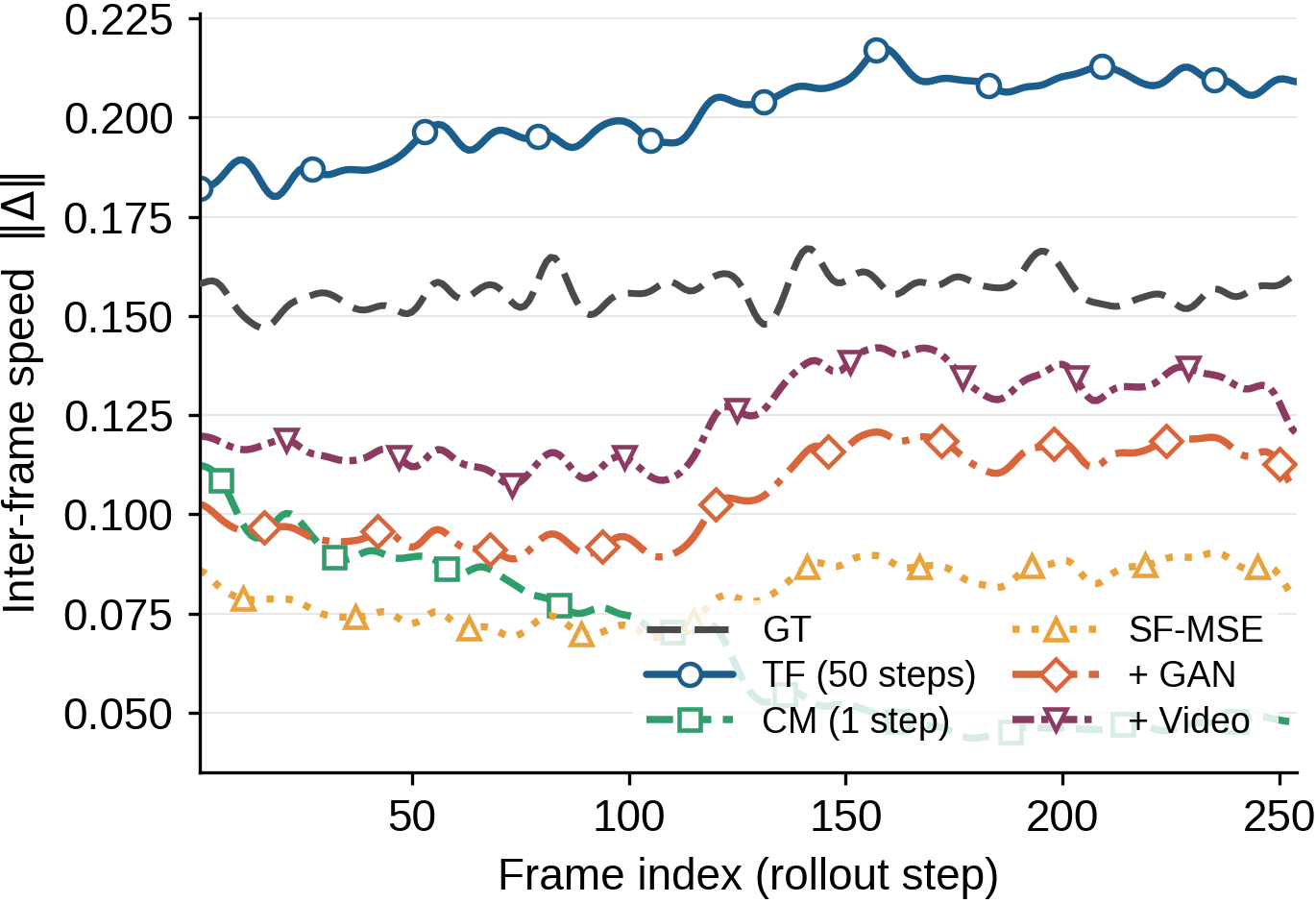}    \caption*{(b) Inter-frame velocity.}
\end{minipage}
\caption{Exposure bias across variants. TF models drift over the rollout (a); SF variants curb this drift, and distribution-matching objectives restore GT-level velocity (b).}
\label{fig:ablation_exposure}
\end{figure}

Figure~\ref{fig:ablation_exposure} isolates exposure bias directly. TF models accumulate error rapidly along the rollout while SF variants grow far more slowly (a), confirming that self-forcing suppresses drift. Velocity is complementary (b): SF-MSE falls well below GT as its mean-seeking objective flattens dynamics, whereas $\mathcal{L}_{\text{adv}}$ and $\mathcal{L}_{\text{DMD}}^{\text{motion}}$ each pull it back toward GT. The two TF models fail in opposite ways: the 50-step teacher overshoots GT velocity, its drift surfacing as erratic oversized motion; the 1-step distilled model collapses to near-static, as distillation strips the sampling stochasticity that gave the teacher its dynamics, leaving exposure bias flatten.

\noindent\textbf{Hierarchical conditioning.}
\label{sec:ablation-cond}
We ablate the two conditioning pathways of $\mathcal{F}_{\text{motion}}$ in Table~\ref{tab:ablation-cond}.

\begin{table}[ht]
\centering
\setlength{\tabcolsep}{4pt}
\begin{tabular}{@{}l|ccccc@{}}
\toprule
\textbf{Variant} & \textbf{MSE}$\downarrow$ & \textbf{CosSim}$\uparrow$ & \textbf{FMD}$\downarrow$ & \textbf{Std-R}$\uparrow$ & \textbf{VC-R}$\uparrow$ \\
\midrule
w/o Global    & 0.314 & 0.807 & 35.08 & 0.483 & 0.443 \\
w/o Local     & 0.588 & 0.725 & 50.44 & 0.578 & 0.282 \\
Full          & \textbf{0.298} & \textbf{0.826} & \textbf{17.07} & \textbf{0.704} & \textbf{0.644} \\
\bottomrule

\end{tabular}
\caption{Ablation on hierarchical conditioning.}
\label{tab:ablation-cond}
\end{table}

The two pathways fail in near-orthogonal ways. Removing local audio conditioning nearly doubles MSE and sharply cuts velocity while barely touching amplitude, consistent with its role in frame-level lip alignment rather than motion scale. Removing the global Q-Former branch leaves MSE comparatively intact but collapses amplitude, consistent with its role as a sequence-level prior on motion intensity rather than fine-grained details. Both ablations more than double FMD, showing that either pathway alone is insufficient for distributional fidelity even when per-frame error looks acceptable, and the two branches are complementary.

\section{Conclusion}
We presented Motar, a streaming talking-head framework that fuses audio and text in an identity-disentangled motion space rather than the video latent, so that conditions control the video transitively and high fidelity no longer requires a large backbone. A hierarchical conditioning scheme routes the two modalities by their temporal granularity, using a global branch for sequence-level emotion and amplitude and a windowed local branch for frame-level lip articulation, so that a motion generator of only 77M parameters achieves precise, controllable synthesis. To make the decomposition stream, our decoupled self-forcing distillation resolves the exposure bias of both the autoregressive motion generator and the renderer under a single frozen teacher, lifting the fidelity ceiling from the motion generator onto the far stronger renderer. The two models run as parallel causal streams, reaching high throughput and low latency.

\newpage

\bibliography{ref}

\end{document}